# Inverse Kinematics, Identification, RIC-based Control, and implementation of an Aerial Manipulator

Ahmed Khalifa, Mohamed Fanni

Department of Mechatronics and Robotics Engineering
Egypt-Japan University of Science and Technology, New Borg-El-Arab city, Alexandria, Egypt
ahmed.khalifa@ejust.edu.eg, mohamed.fanni@ejust.edu.eg,

*Abstract*

This paper presents the inverse kinematic analysis and parameters identification of a novel aerial manipulation system. This system consists of 2-link manipulator attached to the bottom of a quadrotor. This new system presents a solution for the limitations found in the current quadrotor manipulation system. By deriving the inverse kinematics, one can design the controller such that the desired end effector position and orientation can be tracked. To study the feasibility of the proposed system, a quadrotor with high enough payload to add the 2-link manipulator is designed and constructed. Experimental setup of the system is introduced with an experiment to estimate the rotors parameters. Its parameters are identified to be used in the simulation and controller design of the proposed system. System dynamics are derived briefly based on Newton Euler Method. The controller of the proposed system is designed based on Robust Internal-loop Compensator (RIC) and compared to Fuzzy Model Reference Learning Control (FMRLC) technique which was previously designed and tested for the proposed system. These controllers are tested for provide system stability and trajectory tracking under the effect of picking as well as placing a payload and under the effect of changing the operating region. Simulation framework is implemented in MATLAB/SIMULINK environment. The simulation results indicate the effectiveness of the inverse kinematic analysis and the proposed control technique.

## 1 Introduction

Quadrotor is one of the Unmanned Aerial Vehicles (UAVs) which offer possibilities of speed and access to regions that are otherwise inaccessible to ground robotic vehicles. Quadrotor vehicles possess certain essential characteristics, such as small size and cost, Vertical Take Off and Landing (VTOL), performing slow precise movements, and impressive maneuverability, which highlight their potential for use in vital applications. Such applications include; homeland security (e.g. Border patrol and surveillance), and earth sciences (to study climate change, glacier dynamics, and volcanic activity) [23], [4], [11], and [1]. However, most research on UAVs has typically been limited to monitoring and surveillance applications where the objectives are limited to "look" and "search" but "do not touch". Due to their superior mobility, much interest is given to utilize them for mobile manipulation such as inspection of hard-to-reach structures or transportation in remote areas. Previous research on aerial manipulation can be divided into three categories. The first approach is to install a gripper at the bottom of an UAV to hold a payload. In [2], [18], and [28], a quadrotor with a gripper is used for transporting blocks and to build structures. The second approach is to suspend a payload with cables. In [10], an adaptive controller is presented to avoid swing excitation of a payload. In [20], specific attitude and position of a payload is achieved using cables connected to three quadrotors. The other types of research are concerned about interaction with existing structures, as example, for contact inspection. In [27] and [7] research has been conducted on utilizing a force sensor or a brush as a manipulator. However, the above approaches have limitations for manipulation.

For the first category using a gripper, payloads are rigidly connected to the body of an UAV. Accordingly, not only the attitude of the payload is restricted to the attitude of the UAV, but also the accessible range of the end effector is confined because of the UAV body and blades. In the second type using cables, the movement of the payload cannot be always regulated directly because manipulation is achieved using a cable which cannot always drive the motion of the payload as desired. The last cases are applicable to specialized missions such as wall inspection or applying normal force to a surface.

To overcome these limitations, one alternative approach is to equip an aerial vehicle with a robotic manipulator that can actively interact with the environment. For example, in [17], a test bed including four-DOF robot arms and a crane emulating an aerial robot is proposed. By combining the mobility of the aerial vehicle with the versatility of a robotic manipulator, the utility of mobile manipulation can be maximized. When employing the robotic manipulator, the dynamics of the robotic manipulator is highly coupled with of the aerial vehicle, which should be carefully considered in the controller design for the aerial vehicle. Also, an aerial robot needs to tolerate the reaction forces from the interactions with the object or external environment. These reaction forces may affect the stability of an aerial vehicle significantly.

In [21], we propose a new aerial manipulation system that consists of a 2-link manipulator attached to the bottom of a quadrotor. This new system presents a solution for the limitations found in the current quadrotor manipulation system. It has the capability of manipulating the objects with arbitrary location and orientation (DOF are increased from 4 to 6), the manipulator provides sufficient distance between quadrotor and object location, and it is considered as the minimum manipulator weight for aerial manipulation. In [15], The dynamic model of this system is derived taking into account the effect of adding a payload to the manipulator, in addition to, the design of two controllers namely, Direct Fuzzy Logic controller and Fuzzy Model Reference Learning Control applied to this system, are presented. The simulation results indicate the outstanding performance of the FMRLC and the feasibility of the proposed robot. This proposed system opens new application area for robotics. Such applications are inspection, maintenance, firefighting, service robot in crowded cities to deliver light stuff such as post mails or quick meals, rescue operation, surveillance, demining, performing tasks in dangerous places, or transportation in remote places.

In [22], a quadrotor with light-weight manipulators are tested, although the movement of manipulator is not explicitly considered during the design of the PID controller. In [16], an aerial manipulation using a quadrotor with a 2 DOF robotic arm is presented but with different configuration from us. It did not provide a solution for the limited DOFs problem of aerial manipulation, in addition to, it did not provide explicit solution to the inverse kinematics problem.

In this paper the design, kinematic (forward and inverse) and dynamic analysis (including effect of adding a payload to the manipulator end effector), experiment to identify rotors parameters, and control of the proposed quadrotor manipulation system based on RIC, are presented.

This paper is organized as following. Design of the proposed system is described in section 3. Section 4 introduces the system kinematic and dynamic analysis. The rotors parameters identification experiment is described in section 5. The proposed control system is presented in section 6. In section 7, simulation results using MATLAB/SIMULINK are presented. Finally, the main contributions are concluded in section 8.

## 2  Design of the Proposed System

The structure of the proposed system is shown in Fig. 1. The proposed quadrotor manipulation system consists mainly from two parts; the quadrotor and the manipulator.

### 2.1  The Two-Link Manipulator

The design of this manipulator is based on light weight and enough workspace under the quadrotor.

Our target is to design a light and simple 2 DOF manipulator that can carry as much as possible of a payload. One of the available and famous company to sell the components of such type of manipulator is "lynxmotion" [19]. The arm components are selected, purchased and assembled such that the total weight of arm is 200 g and can carry a payload of 200 g [14]. The arm components are:

- Three servo motors: HS-422 for gripper, HS-5485HB for joint 1, and HS-422 for joint 2.
- Serial servo controller (SSC-32): Interface between the main control unit and the servo motors.
- Arduino board (Mega 2560) [8]: Implement manipulator control algorithm.
- PS2 R/C: Remote controller to send commands to manipulator.
- Motor accessories: Aluminum Tubing - 1.50 in, Aluminum Multi-Purpose Servo Bracket Two Pack, Aluminum Tubing Connector Hub, and Aluminum Long "C" Servo Bracket with Ball Bearings Two Pack.

### 2.2  Quadrotor

The quadrotor components are selected such that it can carry payload = 500 g (larger than the total arm weight including the maximum payload value). Asctec pelican quadrotor [9] is used as the quadrotor platform with the following specification: Autopilot sensor board -

Magnetometer - GPS receiver - Futuba R/C - X-bee- 11.1V LiPo battery - 1.6 GHz Intel Atom processor board - wireless LAN access point.

Fig. 1 shows the experimental setup of the aerial manipulation system, while Fig. 2 shows the whole connected system.

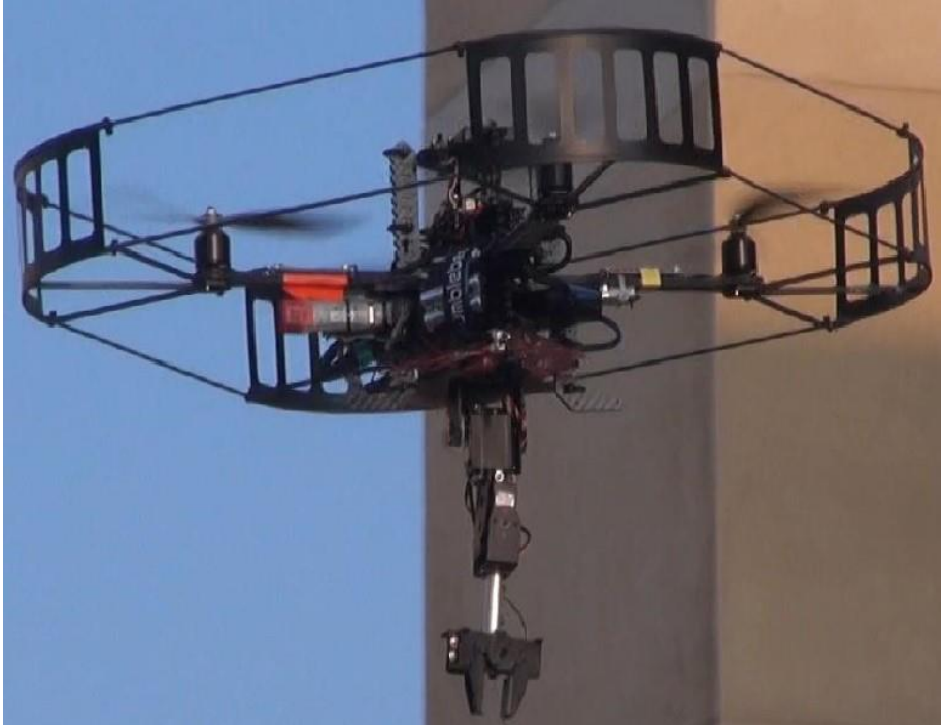

Figure 1: Experimental Setup of the Proposed System

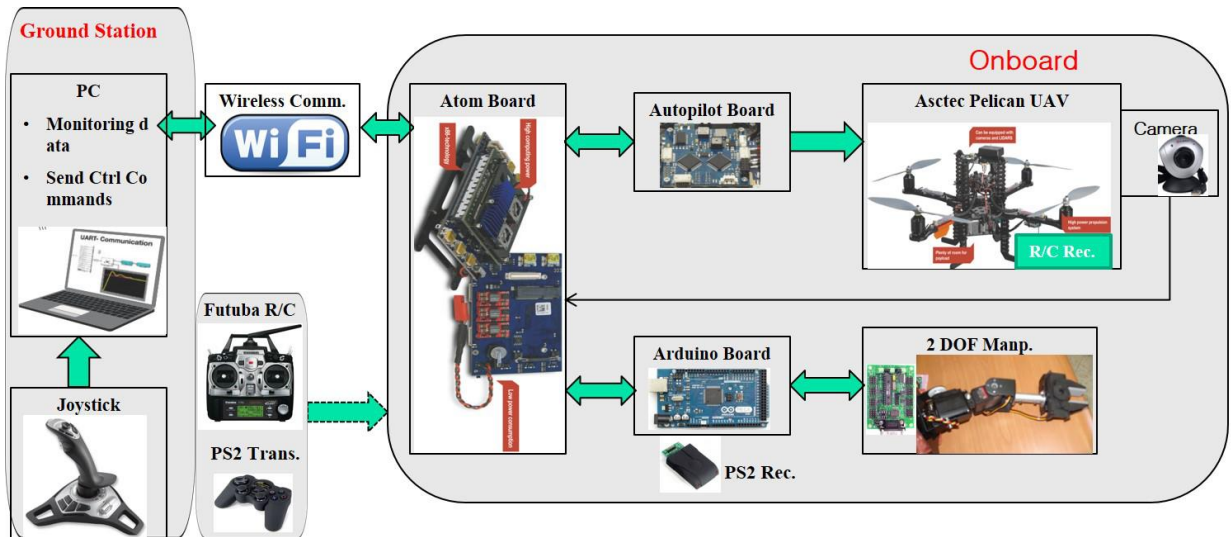

Figure 2: Aerial Manipulation Functional Block Diagram

## 3 Kinematic and Dynamic Analysis

Fig. 3 presents a sketch of the Quadrotor-Manipulator System with the relevant frames. The frames are assumed to satisfy the Denavit-Hartenberg (DH) convention [5].

The manipulator has two revolute joints. The axis of the first revolute joint ($z_0$), that is fixed with respect to the quadrotor, is parallel to the body $x$-axis of the quadrotor (see Fig. 3). The axis of the second joint ($z_1$) will be parallel to the body $y$-axis of quadrotor at home (extended) configuration. Thus, the pitching and rolling rotation of the end effector is now possible independently on the horizontal motion of the quadrotor. Hence, With this new system, the capability of manipulating objects with arbitrary location and orientation is achieved because the DOF are increased from 4 to 6.

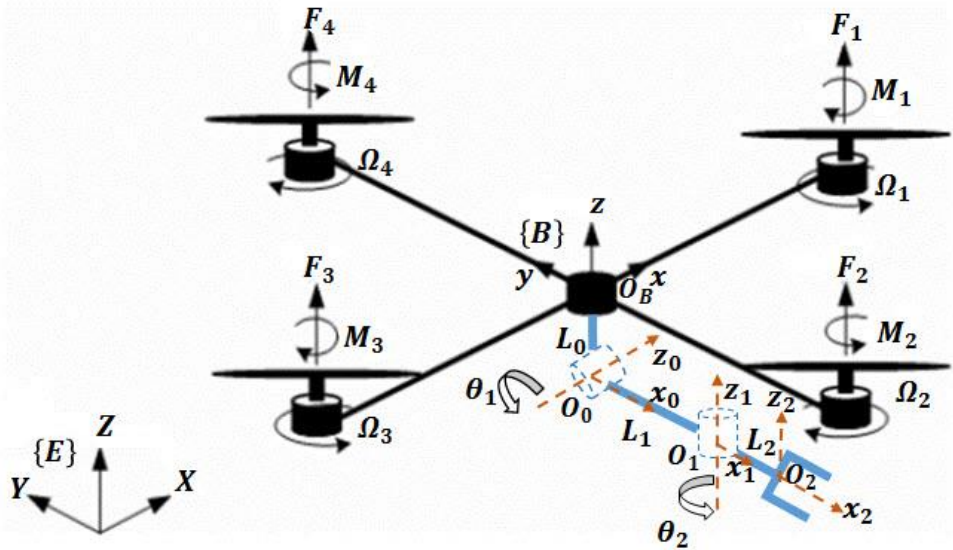

Figure 3: Schematic of Quadrotor Manipulation System Frames

### 3.1 Kinematics

The rotational kinematics of the quadrotor is represented through Euler angles. A rigid body is completely described by its position and orientation with respect to reference frame $\{E\}$, $O_I$-$X\ Y\ Z$, that it is supposed to be earth-fixed and inertial. Let define $\eta_1$ as

$$\eta_1 = [X, Y, Z]^T \tag{1}$$

the vector of the body position coordinates in the earth-fixed reference frame. The vector $\dot{\eta}_1$ is the corresponding time derivative. If one defines

$$v_1 = [u, v, w]^T \tag{2}$$

as the linear velocity of the origin of the body-fixed frame $\{B\}$, $O_B$-$x\ y\ z$, whose origin is coincident with the center of mass ($CM$), with respect to the origin of the earth-fixed frame expressed in the body-fixed frame, the following relation between the defined linear velocities holds:

$$v_1 = R_I^B \dot{\eta}_1 \tag{3}$$

where $R_I^B$ is the rotation matrix expressing the transformation from the inertial frame to the body-fixed frame.

Let define $\eta_2$ as

$$\eta_2 = [\phi, \theta, \psi]^T \qquad (4)$$

the vector of body Euler-angle coordinates in an earth-fixed reference frame. Those are commonly named roll, pitch and yaw angles and corresponds to the elementary rotation around $X$, $Y$ and $Z$ in fixed frame. The vector $\dot{\eta}_2$ is the corresponding time derivative (expressed in the inertial frame). Let define

$$v_2 = [p, q, r]^T \qquad (5)$$

as body-fixed angular velocity. The vector $\dot{\eta}_2$ is related to the body-fixed angular velocity by a proper Jacobian matrix:

$$v_2 = J_v \dot{\eta}_2 \qquad (6)$$

The matrix $J_v$ can be expressed in terms of Euler angles as:

$$J_v = \begin{bmatrix} 1 & 0 & -S(\theta) \\ 0 & C(\phi) & C(\theta)S(\phi) \\ 0 & -S(\theta) & C(\theta)C(\phi) \end{bmatrix} \qquad (7)$$

where $C(\alpha)$ and $S(\alpha)$ are short notations for $cos(\alpha)$ and $sin(\alpha)$. The rotation $R_I^B$ matrix needed to transform the linear velocities, is expressed in terms of Euler angles by the following:

$$R_I^B = \begin{bmatrix} C(\psi)C(\theta) & S(\psi)C(\theta) & -S(\theta) \\ -S(\psi)C(\phi) + S(\psi)S(\theta)C(\psi) & C(\psi)C(\phi) + S(\psi)S(\theta)S(\phi) & C(\theta)S(\phi) \\ S(\psi)S(\phi) + C(\psi)S(\theta)C(\phi) & -C(\psi)S(\phi) + S(\psi)S(\theta)C(\phi) & C(\theta)C(\phi) \end{bmatrix} \qquad (8)$$

The DH parameters for the 2-Link manipulator are derived and presented in [21].

The position and orientation of the end effector relative to the body-fixed frame is easily obtained by multiplying the following homogeneous transformation matrices $A_0^B$, $A_1^0$, $A_2^1$.

### 3.1.1 Forward Kinematics

Let define the position and orientation of the end effector expressed in the inertial frame, as $\eta_{ee_1}$ and $\eta_{ee_2}$ respectively.

$$\eta_{ee_1} = [x_{ee}, y_{ee}, z_{ee}]^T \qquad (9)$$

$$\eta_{ee_2} = [\phi_{ee}, \theta_{ee}, \psi_{ee}]^T \qquad (10)$$

The forward kinematics problem consists of determining the operational coordinates ($\eta_{ee_1}$ and $\eta_{ee_2}$) of the end effector, as a function of the quadrotor movements ($X$, $Y$, $Z$, and $\psi$) as well as the motion of the manipulator's joints ($\theta_1$ and $\theta_2$). This problem is solved by computing the homogeneous transformation matrix composed of relative translations and rotations.

The transformation matrix from the body frame to the inertial frame $A_B^I$ which is:

$$A_B^I = R_B^I * transl(X, Y, Z) \qquad (11)$$

where $R_B^I$ is $4 \times 4$ matrix, and $transl(X,Y,Z)$ is $4 \times 4$ matrix that describes the translation of $X$, $Y$ and $Z$ in the inertial coordinates. The total transformation matrix that relates the end effector frame to the inertial frame is $T_2^I$, which is given by:

$$T_2^I = A_B^I A_0^B A_1^0 A_2^1 \tag{12}$$

Define the general form for this transformation matrix as a function of end effector variables ($\eta_{ee_1}$ and $\eta_{ee_2}$), as following:

$$T_{ee} = \begin{bmatrix} r_{11} & r_{12} & r_{13} & x_{ee} \\ r_{21} & r_{22} & r_{23} & y_{ee} \\ r_{31} & r_{32} & r_{33} & z_{ee} \\ 0 & 0 & 0 & 1 \end{bmatrix} \tag{13}$$

Equating (12) and (13), an expression for the parameters of $T_{ee}$ ($r_{ij}$, $x_{ee}$, $y_{ee}$, and $z_{ee}$; $i,j = 1,2,3$) can be found, from which values of the end effector variables can determined. Euler angles of the end effector ($\phi_{ee}$, $\theta_{ee}$ and $\psi_{ee}$) can be computed from the rotation matrix of $T_{ee}$ as in [13].

### 3.1.2 Inverse Kinematics

The inverse kinematics problem consists of determining the quadrotor movements ($X$, $Y$, $Z$, and $\psi$) as well as the motion of the manipulator's joints ($\theta_1$ and $\theta_2$) as function of operational coordinates ($\eta_{ee_1}$ and $\eta_{ee_2}$) of the end effector.

The inverse kinematics solution is essential for the robot's control, since it allows to compute the required quadrotor movements and manipulator joints angles to move the end effector to a desired position and orientation.

The rotations of the end effector can be parameterized by using several methods one of them, that is chosen, is the euler angles [13].

Equation (12) can be expressed, after putting $\phi = \theta = 0$, since we apply for point-to-point control because we target end effector control during picking and placing positions (reset configuration), as following:

$$T_2^I = \begin{bmatrix} C(\psi)S(\theta_2) + C(\theta_1)C(\theta_2)S(\psi) & C(\psi)C(\theta_2) - C(\theta_1)S(\psi)S(\theta_2) & S(\psi)S(\theta_1) & X + L_1C(\theta_1)S(\psi) + L_2C(\psi \\ S(\psi)S(\theta_2) - C(\psi)C(\theta_1)C(\theta_2) & C(\theta_2)S(\psi) + C(\psi)C(\theta_1)S(\theta_2) & -C(\psi)S(\theta_1) & Y - L_1C(\psi)C(\theta_1) + L_2S(\psi \\ -C(\theta_2)S(\theta_1) & S(\theta_1)S(\theta_2) & C(\theta_1) & Z - L_0 - L_1S(\theta \\ 0 & 0 & 0 & \end{bmatrix}$$

(14)

From (14) and (13), the inverse kinematics of the system can be derived. According to the structure of (14), the inverse orientation is carried out first followed by inverse position. The inverse orientation has three cases as following:

**Case 1**:

Suppose that not both of $r_{13}$, $r_{23}$ are zero. Then from (14), we deduce that $sin(\theta_1) \neq 0$ and $r_{33} \neq \pm 1$. In the same time, $cos(\theta_1) = r_{33}$ and $sin(\theta_1) = \pm\sqrt{1-r_{33}^2}$ and thus,

$$\theta_1 = atan2(\sqrt{1-r_{33}^2}, r_{33}) \tag{15}$$

or

$$\theta_1 = atan2(-\sqrt{1-r_{33}^2}, r_{33}) \tag{16}$$

If we choose the value for $\theta_1$ given by (15), then $sin(\theta_1) > 0$, and

$$\psi = atan2(r_{13}, -r_{23}) \tag{17}$$

$$\theta_2 = atan2(r_{32}, -r_{31}) \tag{18}$$

If we choose the value for $\theta_1$ given by (16), then $sin(\theta_1) < 0$, and

$$\psi = atan2(-r_{13}, r_{23}) \tag{19}$$

$$\theta_2 = atan2(-r_{32}, r_{31}) \tag{20}$$

Thus, there are two solutions depending on the sign chosen for $\theta_1$. If $r_{13} = r_{23} = 0$, then the fact that $T_{ee}$ is orthogonal implies that $r_{33} = \pm 1$.

**Case 2**:

If $r_{13} = r_{23} = 0$ and $r_{33} = 1$, then $cos(\theta_1) = 1$ and $sin(\theta_1) = 0$, so that $\theta_1 = 0$. In this case, the rotation matrix of (14) becomes

$$R_2^I = \begin{bmatrix} S(\theta_2+\psi) & C(\theta_2+\psi) & 0 \\ -C(\theta_2+\psi) & S(\theta_2+\psi) & 0 \\ 0 & 0 & 1 \end{bmatrix} \tag{21}$$

Thus the sum $\theta_2 + \psi$ can be determined as

$$\theta_2 + \psi = atan2(r_{11}, r_{12}) \tag{22}$$

We can assume any value for $\psi$ and get $\theta_2$. Therefor, there are infinity of solutions.

**Case 3**:

If $r_{13} = r_{23} = 0$ and $r_{33} = -1$, then $cos(\theta_1) = -1$ and $sin(\theta_1) = 0$, so that $\theta_1 = \pi$. In this case, the rotation matrix of (14) becomes:

$$R_2^I = \begin{bmatrix} S(\theta_2-\psi) & C(\theta_2-\psi) & 0 \\ C(\theta_2-\psi) & -S(\theta_2-\psi) & 0 \\ 0 & 0 & -1 \end{bmatrix} \tag{23}$$

Thus, $\theta_2 - \psi$ can be determined as

$$\theta_2 - \psi = atan2(r_{11}, r_{12}) \tag{24}$$

One can assume any value for $\psi$ and get $\theta_2$. Therefor, there are infinity of solutions. In cases 2 and 3, putting $\psi = 0$ will lead to find $\theta_2$.

Finally, the inverse position is determined from:

$$X = x_{ee} - (L_1 C(\theta_1)S(\psi) + L_2 C(\psi)S(\theta_2) + L_2 C(\theta_1)C(\theta_2)S(\psi)) \tag{25}$$

$$Y = y_{ee} - (-L_1 C(\psi)C(\theta_1) + L_2 S(\psi)S(\theta_2) - L_2 C(\psi)C(\theta_1)C(\theta_2)) \tag{26}$$

$$Z = z_{ee} - (-L_0 - L_1 S(\theta_1) - L_2 C(\theta_2)S(\theta_1)) \tag{27}$$

### 3.2 Dynamics

The equations of motion of the proposed robot are derived in details in [21]. Applying Newton Euler algorithm [6] to the manipulator considering that the link (with length $L_0$) that is fixed to the quadrotor is the base link, one can get the equations of motion of the manipulator as well as the interaction forces and moments between the manipulator and the quadrotor. The effect of adding a payload to the manipulator will appear in the parameters of its end link, link 2, (e.g. mass, center of gravity, and inertia matrix). Therefore, the payload will change the overall system dynamics.

The equations of motion of the manipulator are:

$$M_1 \ddot{\theta}_1 = T_{m_1} + N_1 \tag{28}$$

$$M_2 \ddot{\theta}_2 = T_{m_2} + N_2 \tag{29}$$

where, $T_{m_1}$ and $T_{m_2}$ are the manipulator actuators' torques. $M_1$, $M_2$, $N_1$, and $N_2$ are nonlinear terms and they are functions in the system states as described in [21].

The Newton Euler method are used to find the equations of motion of the quadrotor after adding the forces/moments from the manipulator are:

$$m\ddot{X} = T(C(\psi)S(\theta)C(\phi) + S(\psi)S(\phi)) + F^I_{m,q_x} \tag{30}$$

$$m\ddot{Y} = T(S(\psi)S(\theta)C(\phi) - C(\psi)S(\phi)) + F^I_{m,q_y} \tag{31}$$

$$m\ddot{Z} = -mg + TC(\theta)C(\phi) + F^I_{m,q_z} \tag{32}$$

$$I_x \ddot{\phi} = \dot{\theta}\dot{\phi}(I_y - I_z) - I_r \dot{\theta}\overline{\Omega} + T_{a_1} + M^B_{m,q_\phi} \tag{33}$$

$$I_y \ddot{\theta} = \dot{\psi}\dot{\phi}(I_z - I_x) + I_r \dot{\phi}\overline{\Omega} + T_{a_2} + M^B_{m,q_\theta} \tag{34}$$

$$I_z\ddot{\psi} = \dot{\theta}\dot{\phi}(I_x - I_y) + T_{a_3} + M^B_{m,q_\psi} \tag{35}$$

where $F^I_{m,q_x}$, $F^I_{m,q_y}$, and $F^I_{m,q_z}$ are the interaction forces from the manipulator to the quadrotor in $X$, $Y$, and $Z$ directions defined in the inertial frame and $M^B_{m,q_\phi}$, $M^B_{m,q_\theta}$, and $M^B_{m,q_\psi}$ are the interaction moments from the manipulator to the quadrotor around $X$, $Y$, and $Z$ directions defined in the body frame.

The variables in (30-35) are defined as follows: $m$ is the mass of the quadrotor. Each rotor $j$ has angular velocity $\Omega_j$ and it produces thrust force $F_j$ and drag moment $M_j$ which are given by:

$$F_j = K_{F_j}\Omega_j^2 \tag{36}$$

$$M_j = K_{M_j}\Omega_j^2 \tag{37}$$

where $K_{F_j}$ and $K_{M_j}$ are the thrust and drag coefficients.

$T$ is the total thrust applied to the quadrotor from all four rotors, and is given by:

$$T = \sum_{i=1}^{4}(F_j) \tag{38}$$

$T_{a_1}$, $T_{a_2}$, and $T_{a_3}$ are the three input moments about the three body axes, and are given as:

$$T_{a_1} = d(F_4 - F_2) \tag{39}$$

$$T_{a_2} = d(F_3 - F_1) \tag{40}$$

$$T_{a_3} = -M_1 + M_2 - M_3 + M_4 \tag{41}$$

$d$ is the distance between the quadrotor center of mass and rotor rotational axis.

$$\overline{\Omega} = \Omega_1 - \Omega_2 + \Omega_3 - \Omega_4 \tag{42}$$

$I_r$ is the rotor inertia. $I_f$ is the inertia matrix of the vehicle around its body-frame assuming that the vehicle is symmetric about x-, y- and z-axis.

## 4 System Parameters Estimation

In order to test the feasibility of the proposed system, a simulation environment will be built. Thus, there is a need to find the real parameters of the system to make the simulation results more accurate and reliable. The identified parameters include the structure parameters and rotor assembly (Electronic Speed Controller, Brush-less DC Motor, and Propeller) parameters ($K_F$ and $K_M$). These parameters will be used in the system simulation and controller design later. A CAD model is developed using SOLIDWORKS to calculate the mass moments of inertia and all the missing geometrical parameters. Fig. 4 shows the experimental setup of quadrotor to estimate the rotor parameters. In this experiment, the rotor is mounted on a 6-DOF torque/force sensor that is connected to a NI Data Acquisition Card (NI DAC). Then, the DAC is connected to a PC running SIMULINK program as an interface to read data from DAC. The velocity of rotor is changed step by step and each time the generated thrust and drag moment is measured and record using SIMULINK program. By using MATLAB Cure Fitting toolbox the generated date are fitted, thus the thrust and moment coefficients can be obtained.

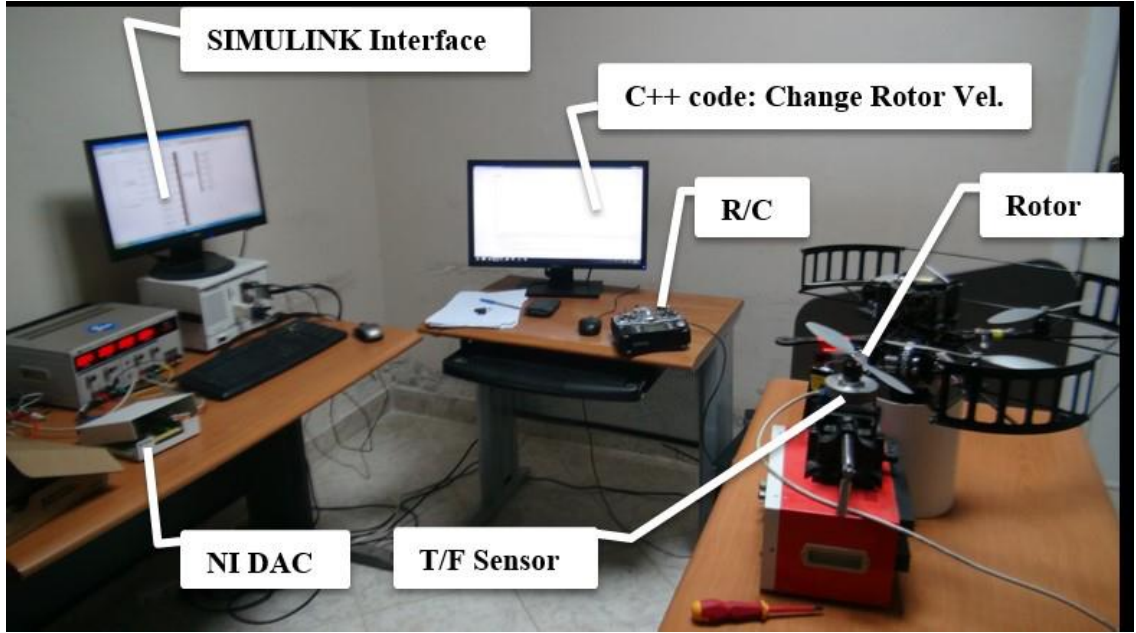

Figure   4: Experiment to estimate rotor coefficients

The identified parameters are given in Table 1.

Table   1: System Parameters

| Par. | Value | Unit | Par. | Value | Unit |
|---|---|---|---|---|---|
| $m$ | 1 | $kg$ | $L_2$ | $85 \times 10^{-3}$ | $m$ |
| $d$ | $223.5 \times 10^{-3}$ | $m$ | $m_0$ | $30 \times 10^{-3}$ | $kg$ |
| $I_x$ | $13.215 \times 10^{-3}$ | $N.m.s^2$ | $m_1$ | $55 \times 10^{-3}$ | $kg$ |
| $I_y$ | $12.522 \times 10^{-3}$ | $N.m.s^2$ | $m_2$ | $112 \times 10^{-3}$ | $kg$ |
| $I_z$ | $23.527 \times 10^{-3}$ | $N.m.s^2$ | $I_r$ | $33.216 \times 10^{-6}$ | $N.m.s^2$ |
| $L_0$ | $30 \times 10^{-3}$ | $m$ | $L_1$ | $70 \times 10^{-3}$ | $m$ |
| $K_{F_1}$ | $1.667 \times 10^{-5}$ | $kg.m.rad^{-2}$ | $K_{F_2}$ | $1.285 \times 10^{-5}$ | $kg.m.rad^{-2}$ |
| $K_{F_3}$ | $1.711 \times 10^{-5}$ | $kg.m.rad^{-2}$ | $K_{F_4}$ | $1.556 \times 10^{-5}$ | $kg.m.rad^{-2}$ |

| $K_{M_1}$ | $3.965 \times 10^{-7}$ | $kg.m^2.rad^{-2}$ | $K_{M_2}$ | $2.847 \times 10^{-7}$ | $kg.m^2.rad^{-2}$ |
|---|---|---|---|---|---|
| $K_{M_3}$ | $4.404 \times 10^{-7}$ | $kg.m^2.rad^{-2}$ | $K_{M_4}$ | $3.170 \times 10^{-7}$ | $kg.m^2.rad^{-2}$ |

## 5  Controller Design

Quadrotor is an under-actuated system, because it has four inputs (angular velocities of its four rotors) and six variables to be controlled. By observing the operation of the quadrotor, one can find that the movement in $X$ - direction is based on the pitch rotation, $\theta$. Also the movement in $Y$ - direction is based on the roll rotation, $\phi$. Therefore, motion along $X$ - and $Y$ -axes will be controlled through controlling $\theta$ and $\phi$.

Fig. 5 presents a block diagram of the proposed control system. The desired values for the end effector's position ($x_{ee_d}$, $y_{ee_d}$ and $z_{ee_d}$) and orientation ($\phi_{ee_d}$, $\theta_{ee_d}$ and $\psi_{ee_d}$) are converted to the desired values of the quadrotor ($X_d$, $Y_d$, $Z_d$ and $\psi_d$) and joints variables ($\theta_{1_d}$ and $\theta_{2_d}$) through the inverse kinematics that are derived in section 4. Next, these values is applied to a trajectory generation algorithm which will be explained later. After that, the controller block receives the desired values and the feedback signals from the system and provides the control signals ($T$, $\tau_{a_1}$, $\tau_{a_2}$, $\tau_{a_3}$, $T_{m_1}$ and $T_{m_2}$). The matrix G of the control mixer, in Fig. 5, is used to transform the assigned thrust force and moments of the quadrotor (the control signals) from the controller block into assigned angular velocities of the four rotors. This matrix can be derived from (38-41) and presented as following:

$$\begin{bmatrix} \Omega_1^2 \\ \Omega_2^2 \\ \Omega_3^2 \\ \Omega_4^2 \end{bmatrix} = \underbrace{\begin{bmatrix} K_{F_1} & K_{F_2} & K_{F_3} & K_{F_4} \\ 0 & -dK_{F_2} & 0 & dK_{F_4} \\ -dK_{F_1} & 0 & dK_{F_3} & 0 \\ -K_{M_1} & K_{M_2} & -K_{M_3} & K_{M_4} \end{bmatrix}}_{\tilde{G}}^{-1} \begin{bmatrix} T \\ \tau_{a_1} \\ \tau_{a_2} \\ \tau_{a_3} \end{bmatrix} \quad (43)$$

Finally, The actual values of the quadrotor and joints are converted to the actual values of the end effector variables through the forward kinematics which are derived in section 4.

The control design criteria are to achieve system stability and zero position error, for the movements in $X$, $Y$, $Z$, and $\psi$ directions as well as for joints' angles $\theta_1$ and $\theta_2$ and consequently for the end effector variables ($\eta_{ee_1}$ and $\eta_{ee_2}$), under the effect of:

- Picking and placing a payload.
- Changing the operating region of the system.

Noting that in the task space, a position tracking is implemented, and in the joint space, trajectory tracking is required.

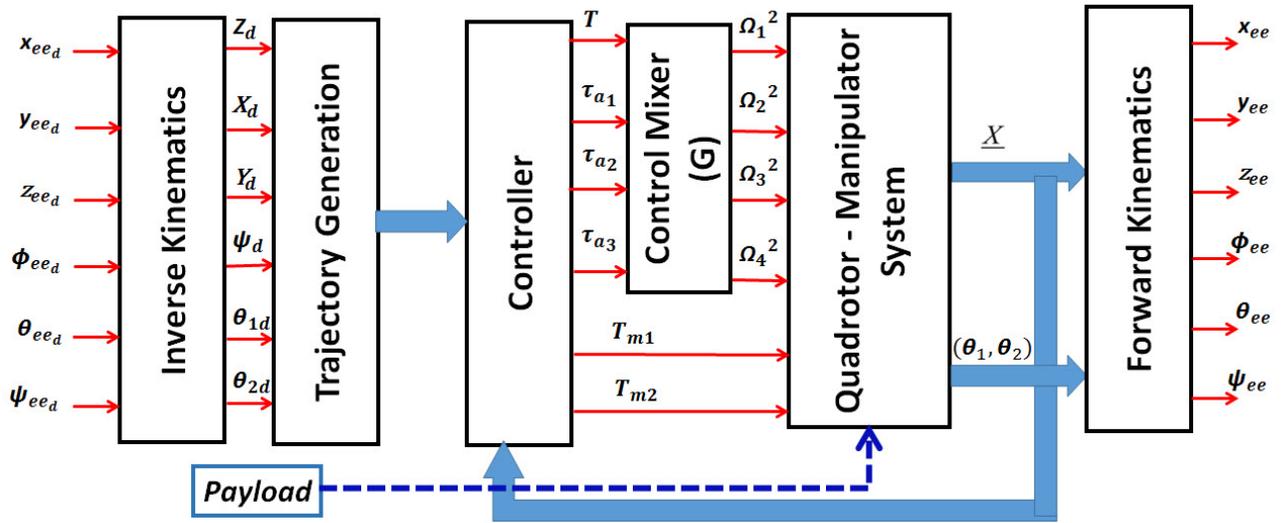

Figure 5: Block Diagram of the Control System

### 5.1 Robust Internal-loop Compensator Based Control

Disturbance-observer (DOB)-based controller design is one of the most popular methods in the field of motion control. In [3], the (DOB)-based controller is designed to realize a nominal system which can control acceleration in order to realize fast and precise servo system, even if servo system has parameter variation and suffers from disturbance. In [12, 24], the generalized disturbance compensation framework, named the robust internal-loop compensator (RIC) is introduced and an advanced design method of a DOB is proposed based on the RIC. In [25], the developed quadrotor shows stable flying performances under the adoption of RIC based disturbance compensation. Although a model is incorrect, RIC method can design a controller by regarding the inaccurate part of the model and sensor noises as disturbances.

We propose a robust internal loop compensator based control as robust controller to get accurate positioning of the proposed system. The controller consists of two parts, internal and external loop. Internal loop is used as a compensator for canceling disturbances, uncertainties and nonlinearities including difference between reference model and real system, and external loop is designed to meet the specification of the system using the result of internal loop compensator.

The RIC based control algorithm, as shown in Fig. 6, controls the response of the plant $P(s)$ to follow that of the model plant $P_m(s)$ even though disturbances $d_{ex}$ and sensor noise $\zeta$ are applied to the plant [12, 25, 26]. RIC based disturbance compensator can be used for position, attitude, and manipulator's joints control in the same way.

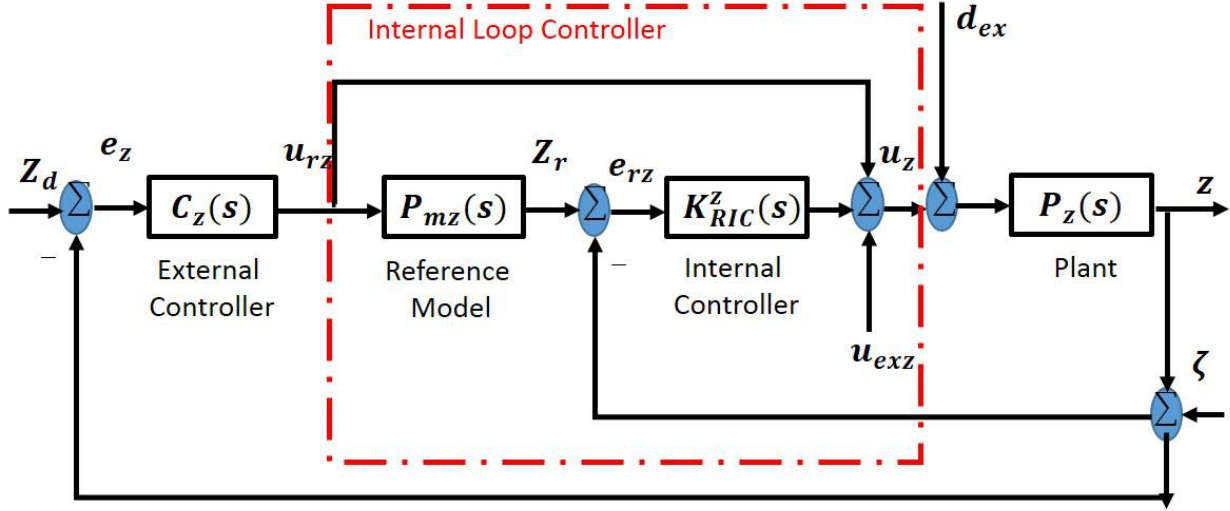

Figure 6: RIC Disturbance Compensation Controller

For all controllers, the reference plant model are given in the form of:

$$P_{mi}(s) = \frac{1}{\tau_{c_i} s^2} \tag{44}$$

where $y_m(s)$ is the output response of the reference model (nominal plant), and $y_r(s)$ is the desired value of the plant. The value $\tau_i$ ($i = x, y, z, \phi, \theta, \psi, \theta_1$, and $\theta_2$), which depends on the plant dynamics, is mass for $x, y, z$-controller and mass moment of inertia for $\phi, \theta, \psi, \theta_1$, and $\theta_2$.

The external-loop compensator $C_z(s)$ for altitude ($z$) control, for instance, are given like PD controller as follows:

$$C_z(s) = k_{pz} + k_{dz} s \tag{45}$$

with the error $e_z = z_r - z$ as the controller input. where $k_{pz}$ and $k_{dz}$ are $P$- and $D$-gain of the external-loop compensator, respectively. The output of the external-loop compensator, i.e., the reference input of RIC is given as

$$u_{rz}(s) = C_z(s) e_z \tag{46}$$

The output of the reference model is compared to the actual response generating the reference error $e_{rz} = z_r - z$ which is applied to internal controller $K_{RIC}^z(s)$ that is chosen to be a PID-like controller and it is given as follows:

$$K_{RIC}^z(s) = k_p^z + k_d^z s + k_i^z \frac{1}{s} \tag{47}$$

Thus, the final control signal $u_z$ is given as:

$$u_z = u_{cz} + u_{kz} + u_{exz} \tag{48}$$

where $u_{cz}$ and $u_{kz}$ are the control signals from the external and internal controllers respectively, while $u_{exz}$ is an external value equal to the robot weight to compensate system weight ($mg$). The procedures for obtaining the RIC control input for $X$, $Y$, $\phi$, $\theta$, $\psi$, $\theta_1$, and $\theta_2$ control are the same with that for altitude ($Z$) control except that $u_{exz}$ equal 0. In addition, there is difference in the design of $X$ and $Y$ controllers. In this control strategy, the desired pitch and roll angles, $\theta_d$ and $\phi_d$, are not explicitly provided to the controller. Instead, they are continuously calculated by $X$ and $Y$ controllers in such a way that they stabilize the quadrotor's attitude. However, there is a need to convert the error and its rate of $X$ and $Y$ that is defined in the inertial frame into their corresponding values defined in the body frame. This conversion is done using the transformation matrix, defined in (8), assuming small angles ($\phi$ and $\theta$) as following:

$$\tilde{x} = \tilde{X}\cos(\psi) + \tilde{Y}\sin(\psi) \tag{49}$$

$$\tilde{y} = \tilde{X}\sin(\psi) - \tilde{Y}\cos(\psi) \tag{50}$$

## 6   Simulation Results

Quintic Polynomial trajectories [5] are used as the reference trajectories for $X$, $Y$, $Z$, $\psi$, $\theta_1$, and $\theta_2$. Those types of trajectories have sinusoidal acceleration which is better in order to avoid vibrational modes. The desired values of end effector position and orientation (Multi-region of operation and point-to-point control) are used to generate the desired trajectories for $X$, $Y$, $Z$, $\phi$, $\theta$ and $\psi$ using the inverse kinematics and then the algorithm for generating the trajectories.

The system equations of motion and the control laws for both FMRLC and RIC techniques are simulated using MATLAB/SIMULINK program. The design details, simulation results, and parameters of FMRLC can be found in [15].The controller parameters of the RIC controller are given in Table 2. Those parameters are tuned to get the required system performance. The two controllers are tested to stabilize and track the desired trajectories under the effect of picking a payload of value 150 g at instant 15 s and placing it at instant 65 s. The simulation results of both FMRLC and RIC are presented in Fig. 3. These results show that RIC and FMRLC is able to track the desired trajectories (with different operating regions) before, during picking, holding, and placing the payload, in addition to, the RIC results is better than the FMRLC in disturbance rejection capability. Furthermore, the generated desired trajectories of $\theta$ and $\phi$ from RIC are smooth compared with that from FMRLC which are more oscillatory (see Fig. 2 and Fig. 2). Moreover, since the RIC is simpler than FMRLC, the computation time for control laws of RIC is very small compared to that of FMRLC. Therefore, RIC is recommended to be implemented in experimental work.

Table   2: RIC Parameters

| *Par./Val.* | $X$ | $Y$ | $Z$ | $\phi$ | $\theta$ | $\psi$ | $\theta_1$ | $\theta_2$ |
|---|---|---|---|---|---|---|---|---|
| $k_{pi}$ | 0.3 | 0.3 | 5 | 30 | 30 | 5 | 5 | 5 |
| $k_{di}$ | 0.7 | 0.7 | 3 | 5 | 5 | 3 | 3 | 3 |

| $k_p^i$ | 0.001 | 0.001 | 5 | 30 | 30 | 5 | 5 | 5 |
|---|---|---|---|---|---|---|---|---|
| $k_d^i$ | 0.001 | 0.001 | 3 | 5 | 5 | 3 | 3 | 3 |
| $k_i^i$ | 0 | 0 | 1 | 10 | 10 | 1 | 1 | 1 |
| $\tau_{ci}$ | 1 | 1 | 1 | 0.01 | 0.01 | 0.02 | 0.1 | 0.1 |

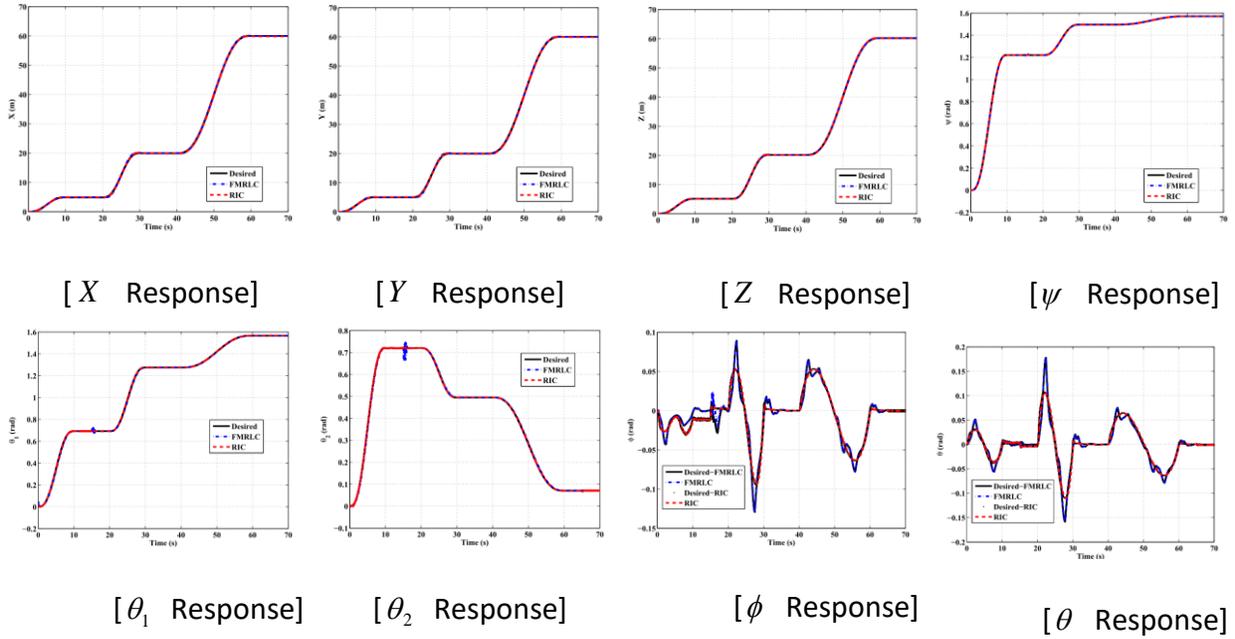

[$X$ Response]    [$Y$ Response]    [$Z$ Response]    [$\psi$ Response]

[$\theta_1$ Response]    [$\theta_2$ Response]    [$\phi$ Response]    [$\theta$ Response]

Figure 3: The Actual Response of RIC and FMRLC Techniques for the Quadrotor and Manipulator Variables: a) $X$, b) $Y$, c) $Z$, d) $\psi$, e) $\theta_1$, f) $\theta_2$, g) $\phi$, and h) $\theta$.

The end effector position and orientation can be found from the forward kinematics (see Fig. 4).

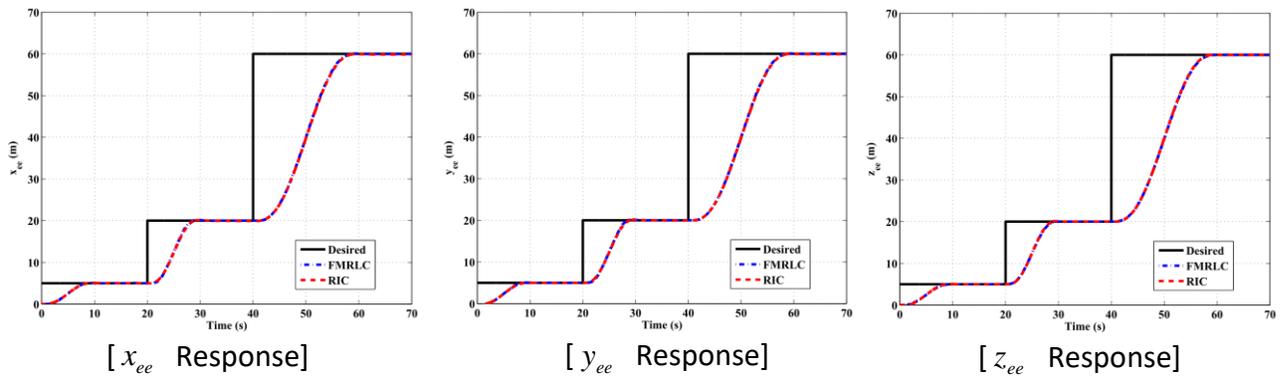

[$x_{ee}$ Response]    [$y_{ee}$ Response]    [$z_{ee}$ Response]

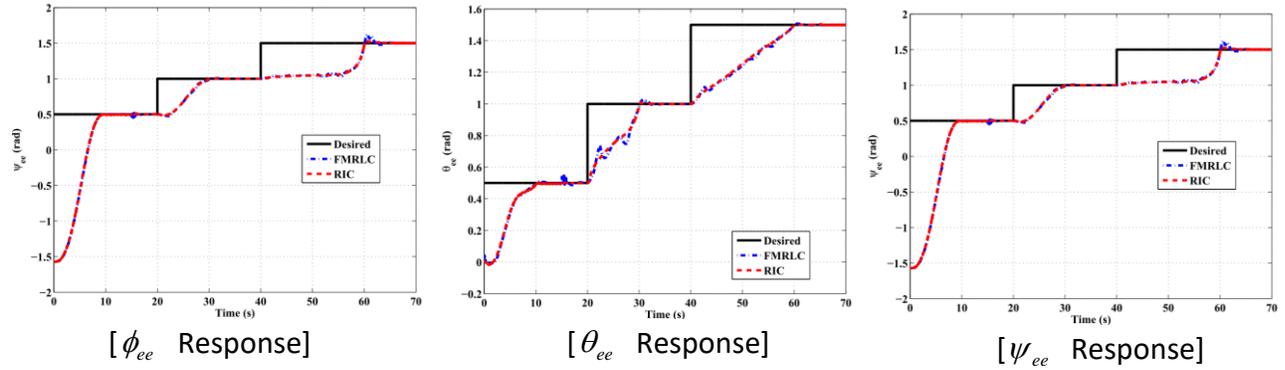

| [$\phi_{ee}$ Response] | [$\theta_{ee}$ Response] | [$\psi_{ee}$ Response] |

Figure 4: The Actual Response of both RIC and FMRLC Techniques for the End Effector Position and Orientation: a) $x_{ee}$, b) $y_{ee}$, c) $z_{ee}$, d) $\phi_{ee}$, e) $\theta_{ee}$, and f) $\psi_{ee}$.

## 7 Conclusion

A new aerial manipulation robot called "Quadrotor Manipulation System" is briefly described. Kinematics, Dynamics and Control of the proposed system are discussed. Experimental setup of the proposed robot is shown and it is used with 6 DOF torque/force sensor to identify the rotor parameters. A closed form for system inverse kinematics which is very simple compared to other trials in this direction. However, this form is used for point-to-point control because we target end effector control during picking and placing positions (reset configuration). RIC based control design is presented to control the proposed system and is compared to the FMRLC. These controllers are tested to provide system stability and trajectory tracking under the effect of picking and placing a payload and the effect of changing the operating region. The system equations of motion are simulated using MATLAB/SIMULINK. Simulation results show that the RIC based control is very simple, has low computation time, and has higher disturbance rejection abilities comparing with FMRLC. In addition, these results indicate the feasibility of the proposed system. Therefore, the RIC is highly recommended to be implemented in real time to experimentally control the proposed system.

## Acknowledgment

The first author is supported by a scholarship from the Mission Department, Ministry of Higher Education of the Government of Egypt which is gratefully acknowledged.

# Authors

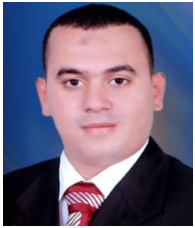

**Ahmed Khalifa** received the B.S. degree in Industrial Electronics and Control Engineering from Monofiya University, Egypt, in 2009, and the M.Sc. degree in Mechatronics and Robotics Engineering from Egypt-Japan University of Science and Technology, Alexandria, Egypt, in 2013. He is currently pursuing the Ph.D. degree in Mechatronics and Robotics Engineering at Egypt-Japan University of Science and Technology.

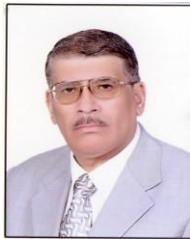

**Mohamed Fanni** received the B.E. and M.Sc. degrees in mechanical engineering from Faculty of Engineering of both Cairo University and Mansoura University, Egypt, in 1981 and 1986, respectively and the Ph.D. degree in engineering from Karlsruhe University, Germany, 1993. He is an Associate Professor with Innovation, Graduate School of Engineering Science, Egypt-Japan University of Science and Technology E-JUST, Alexandria, on leave from Production Engineering & Mechanical Design Department, Faculty of Engineering, Mansoura University, Egypt. His major research interests include robotics engineering, automatic control, and Mechanical Design. His current research focuses on Design & Control of Mechatronic Systems, Surgical Manipulators, Teleoperation systems and Flying/Walking Robots.

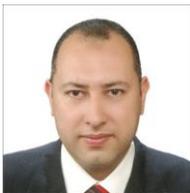

**Ahmed Ramadan** received his B.E. and M. Sc. degrees from Electrical Engineering department in Computer Engineering and Automatic Control from Tanta University, Egypt, in 1996 and 2002 respectively. Then he worked toward the Ph.D. degree from late 2005 till March, 2009 at Systems Innovation Department, Graduate School of Engineering Science, OSAKA University, Japan. Starting from May, 2009 he is a Lecturer / Assistant Professor in the Computer and Control Engineering department in Tanta University. As of the beginning of April 2010, he granted a research fellowship to work as an Assistant Professor in Mechatronics and Robotics Engineering department, Egypt-Japan University and science and technology (EJUST), Egypt. He is a member of the Institute of Electrical and Electronics Engineers (IEEE). His research interests are in the fields of Robotics Engineering, Automatic Control, and Artificial Intelligence Techniques. His current research interests focuses on the design and control of Surgical Manipulators, Master/Slave teleoperation systems, Aerial manipulation and control of Flying Robots.

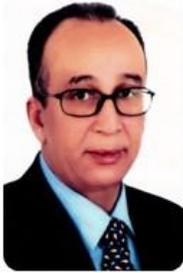 **Ahmed Abo-Ismail** received his Ph.D. degree from Tokyo Institute of Technology, TIT, Japan, 1979. He is a full professor of Automatic Control and Mechatronics, EJUST, Egypt. He is an Honorary Professor of Budapest Tech., Hungary. He is granted a Fulbright award fellowship as a Visiting Professor to PSU, USA, 1987. He is an IFAC Chair (International Federation of Automatic Control) for African, Asian, South American Countries, 2003-2006 and he is a Member of the technical Editorial Board of the ACTA Polytechnica Hungaraica Journal of Applied Sciences, Budapest, Hungary. He is also the Provost, Vice President for Education and Academic Affairs, EJUST, Egypt and former dean of faculty of Engineering, Assuit University, Assuit, Egypt. His major research interests are in the fields of Intelligent and Robust Control, Smart Grippers design, Intelligent Mechatronics System, VR â€"Hepatic Surgical Simulators, Surgical Robots, Assistive Devises for elderly people design and control, Aerial manipulation and control, and Flying/Walking Robots.